# Fuzzy Rankings: Properties and Applications

Jiří Mazurek

*Silesian University in Opava, School of Business Administration in Karviná,*
*Department of Mathematical Methods in Economics.*
*Univerzitní náměstí 1934, Karviná 733 40.*
*e-mail: mazurek@opf.slu.cz.*

**Abstract.** In practice, a ranking of objects with respect to given set of criteria is of considerable importance. However, due to lack of knowledge, information of time pressure, decision makers might not be able to provide a (crisp) ranking of objects from the top to the bottom. Instead, some objects might be ranked equally, or better than other objects only to some degree. In such cases, a generalization of crisp rankings to fuzzy rankings can be more useful. The aim of the article is to introduce the notion of a fuzzy ranking and to discuss its several properties, namely orderings, similarity and indecisiveness. The proposed approach can be used both for group decision making or multiple criteria decision making when uncertainty is involved.

## Introduction

In many areas of human action an ordering of objects from the $1^{st}$ to the $n^{th}$ place (under some criteria) naturally arises. This orderings may express preferences of individuals, groups or a society on a wide range of topics including various products, goods, meals, jobs, services, sportsmen, movies, leisure time activities, etc. In media, rankings of object often appear in a form of various TOP10, TOP100, etc. Ranking of object is also important in a search of a desired object. For a given set of keywords or key phrases, users can search for books, movies, songs or Internet pages. Many Internet stores provide rankings of their goods ordered with respect to price, the year of production, customer recommendations, etc. Rankings are useful when searching for information; the best-known Internet search engine is Google Page rank algorithm which ranks Internet pages according to their relevance for given keywords. Also, the study of rankings or orderings was found extremely important in bioinformatics or genomics.

When considering rankings, one important problem deals with its similarity (dissimilarity), if the same set of objects is ranked by more than one decision maker. Usually, a suitable distance or metric function is employed to assess the similarity, see e.g. Diaconis and Graham (1977), Beck and Lin (1983), Kendall and Gibbons (1990), Cook and Kress (1991), Cook et al. (1996), Fagin et al. (2003), Cook (2006), Tavana et al. (2007), Carterette (2009), Mazurek (2011b) or Farnoud et al. (2014). Other approach takes advantage of some correlation coefficient, such as Kendall's correlation coefficient tau or Spearman's rank correlation coefficient are used, see Spearman (1904) and Kendall (1962).

Another problem stems from the fact that sometimes an expert might hesitate about assigning an object a precise position in a ranking. For instance, a decision maker may say that an object X belongs among three best objects of its category. In such a case, a decision maker is unable to provide a precise ranking of objects, but might be able to provide a fuzzy ranking of objects, that is the ranking where one object can be assigned more than one position. This situation is quite common in a real world due to lack of information, insufficient knowledge of a decision maker, time pressure or conflicting evidence.

Therefore, fuzzy rankings constitute a generalization of (crisp) rankings which might be useful in situations when precise ranking of objects is not possible, or is not suitable. The aim of this paper is to introduce a notion of the fuzzy ranking, and to discuss some of their properties, namely similarity between two fuzzy rankings, ordering of objects and indecisiveness.

The paper is organized as follows: in Section 1 crisp rankings are briefly described, section 2 introduces fuzzy rankings and sections 3-5 provide fuzzy rankings properties, section 6 includes numerical examples and conclusions close the article.

## 1. Crisp rankings

In a crisp ranking each object is assigned its position, so that $n$ objects are ranked from the 1st to the $n^{th}$ place. Thus, every ranking can be considered a permutation of objects.

The permutation of a finite set $A$ is defined as a bijection from $A$ to itself: $\pi : A \to A$. Each permutation (each ranking) of $n$ objects can be represented by a square binary matrix of order $n$ with exactly one value of 1 in each row and column – a *permutation matrix*. For instance, the permutation (ranking) of three objects A, B and C such that A is $1^{st}$, C $2^{nd}$ and B $3^{rd}$, $\pi =$ (A, C, B), can be represented by a matrix $\pi$ with rows corresponding to object and columns to positions:

$$\pi = \begin{pmatrix} 1 & 0 & 0 \\ 0 & 0 & 1 \\ 0 & 1 & 0 \end{pmatrix}$$

Distance-based evaluation of similarity of rankings is based on the premise that the more similar are two ranking, the smaller is their distance. On the other hand, rank correlation coefficients assess similarities (or dissimilarities) in the order of objects listed from the top to the bottom. The advantage of the latter approach rests in the fact that they are normalized to intervals [–1,1]. The higher are the values of a coefficient, the more similar are both rankings.

***Definition 1***. *(Kendall's rank correlation coefficient)*. Let $A$ and $B$ be two rankings on the same domain (on the same set of objects). Kendall's rank correlation coefficient tau is defined as follows (Abdi 2007, Kendall 1962):

$$\tau = \frac{2(n_c - n_d)}{n(n-1)}, \tag{1}$$

where $n_c$ is the number of concordant pairs and $n_d$ is the number of discordant pairs. Concordant (discordant) pair is the ordered pair of objects, which has the same (opposite) order in both rankings. Kendall's tau is normalized to interval $[-1,1]$. In the case of maximum similarity between two rankings $\tau = 1$ (rankings are identical). In the case of maximum dissimilarity $\tau = -1$ (one ranking is reverse of the other).

However, as will be shown in Example 1, rank correlation coefficients alone (or distance-based measures of similarity) are not sufficient in capturing the similarity of rankings, and therefore are not suitable in practice.

***Example 1***. Consider five objects A, B, C, D and E and their rankings $R_1$, $R_2$ and $R_3$ shown in Figure 1. Rankings $R_1$ and $R_2$ differ only by one transposition, namely A and B, for the first and second position. Therefore, from (1) we get: $\tau = \frac{2(9-1)}{5(5-1)} = \frac{16}{20} = 0.8$.

Now consider rankings $R_2$ and $R_3$. Again, only order of one pair of objects, D and E, is reversed, hence $\tau = 0.8$ once again. In terms of Kendall's tau, the similarity of both pairs is the same.

However, in the former case the difference seems to be more important, as the first object (a winner) is different. The change at the 4$^{th}$ or 5$^{th}$ place, in the latter pair, is not so important. Therefore, to evaluate similarity of two rankings, Kendall's distance is not sufficient.

| place | $R_1$ | $R_2$ | $R_3$ |
|---|---|---|---|
| 1 | A | B | A |
| 2 | B | A | B |
| 3 | C | C | C |
| 4 | D | D | E |
| 5 | E | E | D |

**Fig. 1**. Rankings $R_1$, $R_2$ and $R_3$.

To solve the problem of different importance of a position in a ranking, cost or penalty functions were introduced, see e.g. Angelov et al. (2008), Kapah et al. (2009), Kumar and Vassilvistskii (2009) or Farnoud and Milenkovic (2012),

***Definition 2***. Let $n$ be the number of compared objects. Let $P(p_{ij})$ be $n \times n$ symmetric real matrix with entries $p_{ij} \in [0,1]$, $\forall i, j \in \{1,2,...,n\}$, corresponding to interchange of i-th and j-th object in a ranking. Furthermore, $p_{ij} + p_{kl} = p_{il}$, $\forall i, j, k, l \in \{1,2,...,n\}$. Then $P$ is called a penalty matrix.

***Remark 1***. In Definition 2 the constraint $p_{ij} \in [0,1]$ is not necessary, but the normalization of penalty values might be more convenient.

***Example 2***. Figure 2 provides an example of a penalty matrix. For example, the value $p_{12} = 0.5$ means that the transposition of the 1$^{st}$ and 2$^{nd}$ object in a given ranking is penalized by the value 0.5.

$$P = \begin{pmatrix} 0 & 0.5 & 0.8 & 1 \\ 0.5 & 0 & 0.3 & 0.5 \\ 0.8 & 0.3 & 0 & 0.2 \\ 1 & 0.5 & 0.2 & 0 \end{pmatrix}$$

**Fig. 2**. An example of a penalty matrix.

***Definition 3***. Let $A$ be a finite set of objects and let $R_i$ and $R_j$ be rankings of all objects from $A$ expressed in a matrix form. Then, the difference of the two rankings is a matrix given as:

$$D_{ij} = |R_i - R_j|. \tag{2}$$

***Definition 4***. Let $P$ be a penalty matrix and let $D_{ij}$ be a difference of rankings $R_i$ and $R_j$ given by relation (2). Then, the *dissimilarity function DIS of $R_i$ and $R_j$* is given as:

$$DIS(R_i, R_j, P) = \frac{1}{2} \sum_{i=1}^{n} \sum_{j=1}^{n} p_{ij} \cdot d_{ij} . \qquad (3)$$

Maximal dissimilarity $DIS_{max}$ between two rankings $R_i$ and $R_j$ occurs when both rankings are in a reversed order. In such a case, we get:

$$DIS_{max} = \frac{1}{2} \sum_{j>i} p_{ij}, \forall i, j . \qquad (4)$$

**Definition 5**. The similarity *SIM* of two rankings $R_i$ and $R_j$ is given as follows:

$$SIM(R_i, R_j) = 1 - \frac{DIS_{i,j}}{DIS_{max}} . \qquad (5)$$

**Remark 2**. From Definition 3 it is clear that $SIM \in [0,1]$.

## 2. Fuzzy rankings

Fuzzy rankings are generalization of crisp rankings. In crisp rankings, each object was assigned one position, and matrix representation of a crisp ranking is a matrix with only one "1" in each row and column.

However, in real-world decision making problems, a decision maker might be unsure about precise ranking of objects. Consider a situation when four objects, A, B, C and D are ranked so that C is $3^{rd}$ and D is $4^{th}$, but a decision maker is not sure whether A is first or second when compared to B. Therefore, a DM can assign a value 0.5 to $1^{st}$ position for A and 0.5 for $2^{nd}$ position of A, which means a (precise) tie with B. In this case the matrix representation would be that shown in Figure 3a). If a DM is more inclined to A as the $1^{st}$ object, he can assign for example the value 0.7 for A at the $1^{st}$ position and 0.3 for $2^{nd}$ position (which lefts B with 0.3 at $1^{st}$ position and 0.7 at $2^{nd}$ position), see Figure 3b).

This example leads to the notion of a *fuzzy ranking*. Given a set of *n* objects and the set of *n* positions these objects occupy in a ranking, each ordered pair *(object, position)* is assigned the value *f* (a membership degree) from [0,1] interval:

$$f : (object, position) \rightarrow [0,1] . \qquad (6)$$

The value 1 means that a given object belongs to a given position in a given ranking with absolute certainty. On contrary, value 0 expresses that given object does not belong to a given position with absolute certainty.

$$\begin{pmatrix} 0.5 & 0.5 & 0 & 0 \\ 0.5 & 0.5 & 0 & 0 \\ 0 & 0 & 1 & 0 \\ 0 & 0 & 0 & 1 \end{pmatrix} \begin{pmatrix} 0.7 & 0.3 & 0 & 0 \\ 0.3 & 0.7 & 0 & 0 \\ 0 & 0 & 1 & 0 \\ 0 & 0 & 0 & 1 \end{pmatrix}$$

**Fig**. **3a**) and **3b**): Two rankings with uncertain positions.

To guarantee logical consistency (for instance one object cannot be ranked with absolute certainty to two or more different positions), some constraints of the values of the mapping (6) are necessary. This, in turn, leads to the definition of a fuzzy ranking:

**Definition 6**. The fuzzy ranking is every square matrix $R(a_{ij})_{n \times n}$ satisfying the following conditions:

$$\sum_{i=1}^{n} a_{ij} = 1 \text{ and } \sum_{j=1}^{n} a_{ij} = 1. \tag{7}$$

According to Definition 6, every bistochastic matrix can be considered a fuzzy ranking, and vice versa.

The set of permutation matrices $S_n$ of order $n$ is a subset of a set of bistochastic matrices (see Theorem 1 thereinafter). Relations (7) pose $2n - 1$ constraints for $a_{ij}$ values, hence the number of degrees of freedom is $(n - 1)^2$. The set of bistochastic matrices is often denoted as $B_n$, where $B_n$ stands for Birkhoff's polytopes of order $n$.

**Theorem 1** (Birkhoff-von Neumann). *The set of bistochastic matrices of order n is the convex hull[1] of the set of permutation matrices of order n.*

*Proof* : Birkhoff (1948).

With respect to its algebraic structure, the set $S_n$ is a group under matrix multiplication, while the set $B_n$ is only a monoid.

**Proposition 1**. *The set of bistochastic matrices ($B_n$) of order $n \geq 2$ is a monoid under matrix multiplication.*

*Proof*: To be a monoid, the set $B_n$ has to be closed under matrix multiplication, it must satisfy associativity and a unitary element $I$ must exist for all $A \in B_n$. Because matrix multiplication is associative and the identity matrix $I \in B_n$, it suffices to show that $B_n$ is closed under multiplication: Let $A \in B_n, B \in B_n$ and $C = A \cdot B$. Then the sum of all elements in the $k$-th row of the matrix $C$ is: $\sum_{j=1}^{n} a_{kj} \cdot \left( \sum_{i=1}^{n} b_{ji} \right) = \sum_{j=1}^{n} a_{kj} \cdot 1 = 1$, the proof for columns is analogous. As $B_n$ contains singular matrices as well, it is not a group.

**Proposition 2**. *The arithmetic mean and the weighted arithmetic mean (with normalized weights) of fuzzy or crisp permutations is a fuzzy permutation.* (*The set of fuzzy rankings of a given order is closed under arithmetic mean operation*).

*Proof* : obvious from Definition 6.

Proposition 2 states that from an algebraic point of view the set of fuzzy rankings of a given order endowed with the arithmetic mean as a binary operation is only a *magma* (a *grupoid*). However, for practical purposes Proposition 2 ensures closeness: It guarantees the arithmetic mean of fuzzy rankings is a fuzzy ranking again.

---

[1]The convex hull $H(X)$ of a set $X$ (Birkhoff,1948): $H(X) = \left\{ \sum_{i=1}^{n} \theta_i a_i / a_i \in X, \theta_i \in R, \theta_i \geq 0, \sum_{i=1}^{n} \theta_i = 1, n = 1, 2, ... \right\}$.

## 3. Fuzzy rankings and ordering of objects

In the case of crisp rankings, the ranking of all objects is obvious. However, in the case of fuzzy rankings the situation is more complex. To rank all objects of a given fuzzy ranking, the following pair-wise dominance relation is introduced (Mazurek 2011a, 2012):

**Definition 7**. Let $\Pi_{ij}$ be the fuzzy ranking of an object $i$ at a position $j$. Then, a cumulative fuzzy ranking $H_{ij}$ of an object $i$ from the $1^{st}$ to the $j^{th}$ position is given as:

$$H_{ij} = \sum_{k=1}^{j} \Pi_{ik}. \tag{8}$$

**Definition 8**. An object $r$ dominates an object $s$ $(r \succ s)$ if and only if all cumulative fuzzy rankings $H_{rj}$ of an object $r$ are higher (at least once) or equal to cumulative fuzzy rankings $H_{sj}$ of an object $s$:

$$(r \succ s) \Leftrightarrow H_{rj} \geq H_{sj}, \ j \in \{1, 2, ..., n\}. \tag{9}$$

The highest ranking (the best) object is the object which is not dominated by any other object. It should be noted that the dominance relations given in (8) and (9) enable only a partial order of all objects, as some object might be incomparable (tied).

## 4. Fuzzy rankings and similarity

Similarity between two fuzzy rankings $\widetilde{R}_i$ and $\widetilde{R}_j$ can be defined analogously to the similarity of crisp rankings via relations (2-5). Given the two fuzzy rankings of the same set of $n$ objects and corresponding penalty matrix, the similarity of $\widetilde{R}_i$ and $\widetilde{R}_j$ is expressed as a value in the $[0,1]$ interval, when the value 0 expresses minimum similarity and 1 maximum similarity (identity).

**Definition 3´**. Let $A$ be a finite set of objects and let $\widetilde{R}_i$ and $\widetilde{R}_j$ be rankings of all objects from $A$ expressed in a matrix form. Then, the difference $D_{ij}$ of the two rankings is a (crisp) matrix given as:

$$D_{ij} = \left| \widetilde{R}_i - \widetilde{R}_j \right|. \tag{2´}$$

**Definition 4´**. Let $P$ be a penalty matrix and let $\widetilde{D}_{ij}$ be a difference of rankings $\widetilde{R}_i$ and $\widetilde{R}_j$ given by relation (2´). Then, the *dissimilarity function DIS* of $\widetilde{R}_i$ and $\widetilde{R}_j$ is given as:

$$DIS(\widetilde{R}_i, \widetilde{R}_j, P) = \frac{1}{2} \sum_{i=1}^{n} \sum_{j=1}^{n} p_{ij} \cdot d_{ij}. \tag{3´}$$

The maximal dissimilarity $DIS_{max}$ between two rankings $\widetilde{R}_i$ and $\widetilde{R}_j$ is given as:

$$DIS_{max} = \frac{1}{2} \sum_{j>i} p_{ij}, \forall i, j. \tag{4´}$$

**Definition 5´**. The similarity *SIM* of two rankings $\widetilde{R}_i$ and $\widetilde{R}_j$ is given as follows:

$$SIM(\widetilde{R}_i, \widetilde{R}_j) = 1 - \frac{DIS_{i,j}}{DIS_{max}}. \tag{5'}$$

Again, $SIM \in [0,1]$.

## 5. Fuzzy rankings and indecisiveness

Fuzzy rankings framework allows an evaluation of experts' decisions in terms of *indecisiveness*. An expert is absolutely decisive, when he assigns each alternative value 1 for a given position and value 0 to all other positions, and indecisive otherwise. To evaluate indecisiveness, Shannon's entropy as a measure of uncertainty can be used, see e.g. Klir and Folger (1987):

$$H(p(x)) = \sum_{i=1}^{n} -p(x_i) \log_2(p(x_i)). \tag{10}$$

where $p(x_i)$ are probabilities assigned to values $x_i$, $i \in \{1, 2, ..., n\}$; and $p(x) \log_2(p(x)) = 0$ for $p(x) = 0$ by definition.

A decision maker is absolutely indecisive, if he/she provides the fuzzy rankings with the uniform distribution $p(x_i) = \frac{1}{n}$, $i \in \{1, 2, ..., n\}$. In other words, all objects are ranked equally. In this case, the entropy (10) is equal to Hartley's information $I(n)$ (or Hartley's measure of nonspecifity):

$$I(n) = \log_2 n. \tag{11}$$

Therefore, maximum indecisiveness $IND_{max}(n)$ is given as:

$$IND_{max}(n) = n \log_2 n. \tag{12}$$

The indecisiveness *IND* of a fuzzy ranking $\widetilde{R}(a_{ij})$ is given as:

$$IND(\widetilde{R}) = -\sum_{i=1}^{n} \sum_{j=1}^{n} a_{ij} \log_2(a_{ij}). \tag{13}$$

***Definition 9***. Let $\widetilde{R}_i$ be a fuzzy ranking provided by a given decision maker. Then, decision maker's *index of indecisiveness II* is given as follows:

$$II(\widetilde{R}_i, n) = \frac{IND(\widetilde{R}_i)}{n \cdot \log_2 n}. \tag{14}$$

Index of indecisiveness (14) can be used to derive weights of decision makers in a group decision making: experts who provide more decisive ranking (more specific information) might be assigned higher weights.

## 6. Numerical examples

In this section several numerical examples are provided to illustrate the use of fuzzy rankings.

***Example 3***. Consider the fuzzy ranking of 4 objects (A, B, C and D) shown in Figure 4 and order all objects.

Firstly, from relation (8) the cumulative fuzzy rankings $H_{ij}$ of all objects are computed, see Figure 5. Notice from Figure 5, that A is ranked better (higher) than B for the 1$^{st}$ position; A is ranked better than B for the 1$^{st}$ and 2$^{nd}$ position altogether; both objects are equal for the 1$^{st}$, 2$^{nd}$ and 3$^{rd}$ position altogether, and of course, A and B are equal for all positions altogether. Therefore, with the use of relation (9), A dominates B ($A \succ B$). After the evaluation of all pairs, the following overall ranking of all four alternatives is as follows: 1$^{st}$ A, 2$^{nd}$ B, 3$^{rd}$ C and 4$^{th}$ D.

| Object/position | 1. | 2. | 3. | 4. |
|---|---|---|---|---|
| A | 0.30 | 0.5 | 0.20 | 0 |
| B | 0.25 | 0.25 | 0.5 | 0 |
| C | 0.25 | 0.25 | 0 | 0.5 |
| D | 0.25 | 0 | 0.25 | 0.5 |

**Fig. 4**. The fuzzy ranking (permutation) of alternatives A, B, C and D.

| Obejct/position | 1. | 2. | 3. | 4. |
|---|---|---|---|---|
| A | 0.30 | 0.80 | 1 | 1 |
| B | 0.25 | 0.5 | 1 | 1 |
| C | 0.25 | 0.5 | 0.5 | 1 |
| D | 0.25 | 0.25 | 0.5 | 1 |

**Fig. 5.** The cumulative fuzzy ranking ($H_{ij}$) of alternatives A, B, C and D.

***Example 4***. Consider two fuzzy rankings $\widetilde{R}_1$ and $\widetilde{R}_2$ shown in Figures 6 and 7, and the penalty matrix in Figure 2. Then the dissimilarity of both rankings given by relation (3´) is:

$$DIS(\widetilde{R}_1, \widetilde{R}_2, P) = \frac{1}{2} \sum_{i=1}^{n} \sum_{j=1}^{n} p_{ij} \cdot d_{ij} = 0.275$$

$$SIM(\widetilde{R}_1, \widetilde{R}_2) = 1 - \frac{DIS_{1,2}}{DIS_{max}} = 1 - \frac{0.275}{2.3} = 0.880.$$

As for indecisiveness, index of indecisiveness for $\widetilde{R}_1$: II($\widetilde{R}_1$) = 0.801, and II($\widetilde{R}_2$) = 0.871. Therefore, $\widetilde{R}_1$ is less indecisive and provides more specific information than $\widetilde{R}_2$.

| Object/position | 1. | 2. | 3. | 4. |
|---|---|---|---|---|
| A | 0.60 | 0.30 | 0.10 | 0 |
| B | 0.30 | 0.30 | 0.20 | 0.20 |
| C | 0.10 | 0.30 | 0.40 | 0.20 |
| D | 0 | 0.10 | 0.30 | 0.60 |

**Fig. 6**. The fuzzy ranking $\widetilde{R}_1$.

| Object/position | 1. | 2. | 3. | 4. |
|---|---|---|---|---|
| A | 0.40 | 0.30 | 0.20 | 0.10 |
| B | 0.30 | 0.25 | 0.25 | 0.20 |
| C | 0.20 | 0.25 | 0.30 | 0.25 |
| D | 0.10 | 0.20 | 0.25 | 0.45 |

**Fig. 7**. The fuzzy ranking $\widetilde{R}_2$.

| Object/position | 1. | 2. | 3. | 4. |
|---|---|---|---|---|
| A | 0.20 | 0 | 0.10 | 0.10 |
| B | 0 | 0.05 | 0.05 | 0 |
| C | 0.10 | 0.05 | 0.10 | 0.05 |
| D | 0.10 | 0.10 | 0.05 | 0.15 |

**Fig. 8**. The difference $D_{12}$ of rankings $\widetilde{R}_1$ and $\widetilde{R}_2$.

Fuzzy ranking can be also employed in a group decision making framework, where each expert (a decision maker) provides a crisp ranking of (the same set) of given objects. Then, an aggregation of these crisp rankings leads to a fuzzy ranking (see Theorem 1). For aggregation functions and operators see e.g. Grabisch et al. (2009). The same applies to a multiple criteria decision making framework, where crisp rankings of objects with respect to a given set of criteria can be aggregated into one fuzzy ranking.

*Example 5*. Consider 4 decision makers and their crisp rankings of four objects (A, B, C and D) given in Figure 9. Then, with the use of the arithmetic mean as the aggregation operator the fuzzy ranking of the group is achieved, see Fig. 10. At last, all four objects can be ranked with the use of dominance relations (8) and (9):
First we compare A and B (see Figure 11): A is ranked equally with B for the $1^{st}$ position; A is ranked better than B for the $1^{st}$ and $2^{nd}$ position altogether; A is ranked equally with B for the $1^{st}$, $2^{nd}$ and $3^{rd}$ position altogether, which means A dominates B.
By evaluation of all other pairs of objects we get the final (group) ranking of all objects: $1^{st}$ A, $2^{nd}$ B, $3^{rd}$ C and $4^{th}$ D.

$$\begin{array}{cccc|cccc|cccc|cccc}
1 & 0 & 0 & 0 & 0 & 1 & 0 & 0 & 0 & 1 & 0 & 0 & 0 & 0 & 1 & 0 \\
0 & 0 & 1 & 0 & 1 & 0 & 0 & 0 & 0 & 0 & 1 & 0 & 0 & 1 & 0 & 0 \\
0 & 1 & 0 & 0 & 0 & 0 & 1 & 0 & 1 & 0 & 0 & 0 & 0 & 0 & 0 & 1 \\
0 & 0 & 0 & 1 & 0 & 0 & 1 & 0 & 0 & 0 & 0 & 1 & 1 & 0 & 0 & 0
\end{array}$$

**Fig. 9.** DMs' (crisp) rankings of four objects.

| Object/position | 1. | 2. | 3. | 4. |
|---|---|---|---|---|
| A | 0.25 | 0.50 | 0.25 | 0 |
| B | 0.25 | 0.25 | 0.50 | 0 |
| C | 0.25 | 0.25 | 0.25 | 0.25 |
| D | 0.25 | 0 | 0.25 | 0.50 |

**Fig. 10.** Fuzzy (group) ranking of four objects.

| Obejct/position | 1. | 2. | 3. | 4. |
|---|---|---|---|---|
| A | 0.25 | 0.75 | 1 | 1 |
| B | 0.25 | 0.50 | 1 | 1 |
| C | 0.25 | 0.50 | 0.75 | 1 |
| D | 0.25 | 0.25 | 0.5 | 1 |

**Fig. 11.** The cumulative fuzzy ranking ($H_{ij}$) of alternatives A, B, C and D.

# 7. Conclusions

The aim of the article was to introduce fuzzy rankings and some of their properties regarding ordering of objects, a similarity measure between two fuzzy rankings and their indecisiveness. Fuzzy rankings are natural generalization of crisp rankings and may be used in situations when a decision maker is not sure about precise ranking of given objects.

Also, the presented approach can be employed in a group decision making or multiple criteria decision making frameworks, as rankings provided by a set of experts or rankings with respect to given criteria can be easily converted into one fuzzy ranking which is to be further evaluated.